\documentclass{svproc}
\usepackage{url}
\usepackage{graphicx}
\usepackage{subfigure}

\def\degree{${}^{\circ}$}
\makeatletter
\newif\if@restonecol
\makeatother

\usepackage[linesnumbered,ruled,vlined]{algorithm2e}
\usepackage{algpseudocode}
\usepackage{amsmath}
\begin{document}
\mainmatter              
\title{Butterfly detection and classification based on integrated YOLO algorithm}
\titlerunning{Butterfly Detection and Classification Based on Integrated YOLO Algorithm}  
%
\author{Bohan Liang \and Shangxi Wu \and Kaiyuan Xu \and Jingyu Hao}
\authorrunning{Bohan Liang et al.} 
%
\tocauthor{}
\institute{Beijing Jiaotong University, Beijing, China}

\maketitle              

\begin{abstract}
  Insects are abundant species on the earth, and the task of identification
  and identification of insects is complex and arduous. How to apply
  artificial intelligence technology and digital image processing
  methods to automatic identification of insect species is a hot
  issue in current research. In this paper, the problem of automatic
  detection and classification recognition of butterfly photographs is
  studied, and a method of bio-labeling suitable for butterfly
  classification is proposed. On the basis of YOLO algorithm\cite{r1},
  by synthesizing the results of YOLO models with different
  training mechanisms, a butterfly automatic detection and
  classification recognition algorithm based on YOLO algorithm is
  proposed. It greatly improves the generalization ability of Yolo
  algorithm and makes it have better ability to solve small sample
  problems. The experimental results show that the proposed annotation
  method and integrated YOLO algorithm have high accuracy and recognition
  rate in butterfly automatic detection and recognition.
\end{abstract}
\section{Introduction}

As one of the important environmental indicators insects, butterfly species are complex and diverse. The identification of butterfly species is directly related to the crops eaten by humans and animals. At present, reliable butterfly identification methods widely used are not effective. Artificial identification of butterfly species is not only a huge workload, but also requires Long-term Experience and Knowledge Accumulation. How to recognize insect species automatically is one of the hot topics in the field of computer vision.

Automatic recognition of insect species requires recognition and classification of digital images, and the effect of image classification is closely related to the quality of texture feature extraction. In 2004, Gaston et al.\cite{r2} introduced the application of artificial intelligence technology and digital image processing method in digital image recognition. Since then, many experts and scholars have done a lot of work in this area\cite{r3,r4}. In recent years, with the development of machine learning, researchers have proposed many related application algorithms in butterfly detection and detection. In 2012, Wang et al.\cite{r5} used content-based image retrieval (CBIR) to extract image features of butterflies, such as color, shape and texture, compared different features, feature weights and similarity matching algorithms, and proposed corresponding classification methods. In 2014, Kaya Y et al.\cite{r6} applied Local Binary Patterns (LBP)\cite{r7} and Grey-Level Co-occurrence Matrix (GLCM)\cite{r8} to extract the texture features of butterfly images, then used single hidden layer neural network to classify, and proposed an automatic butterfly species recognition method based on extreme learning machine law. In the same year, Kang S H et al.\cite{r9} proposed an effective recognition scheme based on branch length similarity (BLS) entropy profile using butterfly images observed from different angles as training data of neural network. In 2015, based on the morphological characteristics and texture distribution of butterflies, Li Fan\cite{r10} proposed the corresponding feature extraction and classification decision-making methods by using the gray level co-occurrence matrix (GLCM) features of image blocks and K-Nearest Neighbor classification algorithm. In 2016, Zhou A M et al.\cite{r11} proved that the deep learning model is feasible and has strong generalization ability in automatic recognition of butterfly specimens.

In 2018, Xie Juanying et al.\cite{r12} used the method of butterfly detection and species recognition based on Faster-RCNN\cite{r13}, and expanded the data set by using butterfly photos in natural ecological environment. The algorithm has high positioning accuracy and recognition accuracy for butterfly photos in natural ecological environment. Traditional butterfly recognition algorithms have the following problems:
\begin{itemize}
  \item[1.] In natural ecological photos, butterflies often appear in the form of small targets (the area of butterfly image is too small), traditional butterfly recognition algorithms are often powerless.
  \item[2.] The amount of data needed for training is huge, but it can not find high-quality annotated public data sets.
  \item[3.] There are too few pictures of some rare butterflies in the natural state to be used directly as training sets.
\end{itemize}

YOLO model , proposed by Joe Redmon\cite{r1}, is a well-known end-to-end learning model in the field of target detection. Compared with the two-step model of RCNN\cite{r13} series, YOLO model can execute much faster and avoid background errors, but it has poor positioning accuracy and false positive rate (FPR) of some classifications of single model is high.

In order to improve the efficiency of butterfly recognition, this paper will make full use of the data provided by China Data Mining Competition and Baidu Encyclopedia, establish a butterfly data set containing a large number of butterfly ecological photos, train the model using ecological photos in natural environment, and based on YOLO V3 algorithm, propose an integration algorithm  which can be used to locate and identify butterfly ecological photos.
The main structure of the paper is as follows:

\begin{itemize}
  \item[1.] Data Set, Data Annotation and Data Preprocessing.  A butterfly dataset with 2342 precisely labeled natural environments is established by self-labeling with the butterfly ecological photos provided in document\cite{r14} and the pictures in Baidu and other Internet photo databases. A set of image labeling methods suitable for labeling butterflies is sorted out through experiments, and the labeling methods are generalized.
  \item[2.] Integrated YOLO algorithm. This algorithm inherits the recognition speed of YOLO algorithm, optimizes YOLO algorithm, improves the recognition ability of YOLO algorithm in fine target, and the learning ability and generalization ability of YOLO algorithm for small sample. It provides a good idea for subsequent target detection based on atlas.
  \item[3.] Experiments and analysis. The performance of YOLO algorithm in different annotation and processing modes and integrated YOLO algorithm on butterfly test set are presented and analyzed.
  \item[4.] Summary and Prospect.
\end{itemize}

\section{Data Set, Data Annotation and Data Preprocessing}

\subsection{Data Set}

\begin{figure}[htbp]
  \centering
  \subfigure{
  \includegraphics[height=2.2cm]{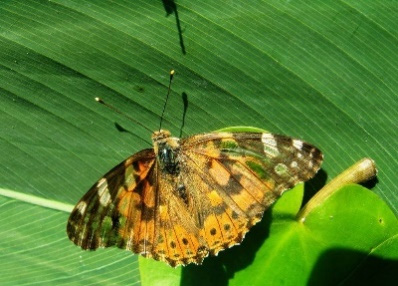}
  }
  \quad
  \subfigure{
  \includegraphics[height=2.2cm]{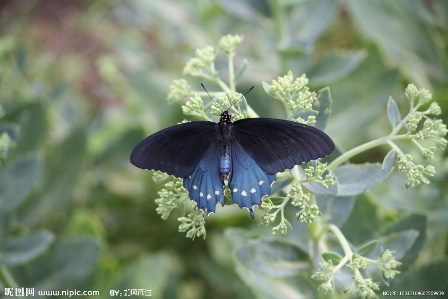}
  }
  \quad
  \subfigure{
  \includegraphics[height=2.2cm]{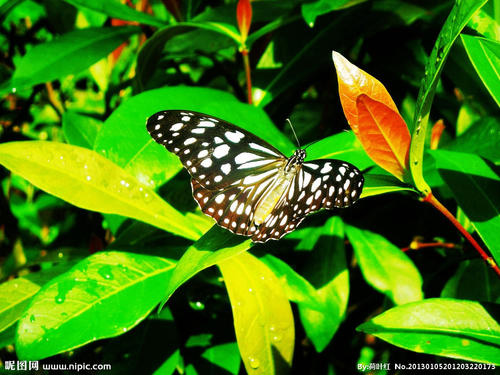}
  }
  
  \subfigure{
  \includegraphics[height=2.2cm]{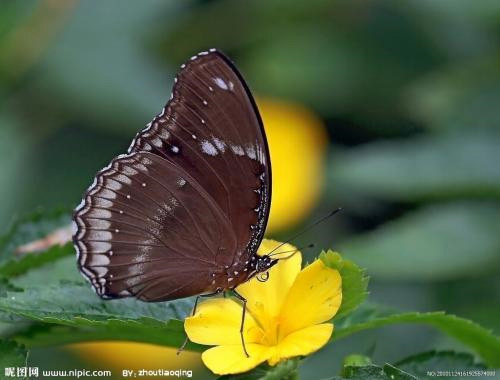}
  }
  \quad
  \subfigure{
  \includegraphics[height=2.2cm]{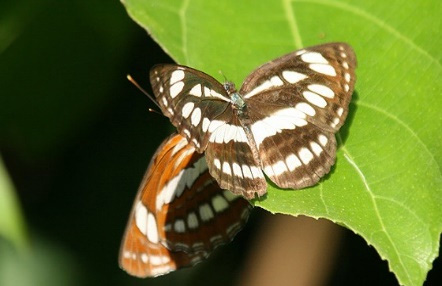}
  }
  \quad
  \subfigure{
  \includegraphics[height=2.2cm]{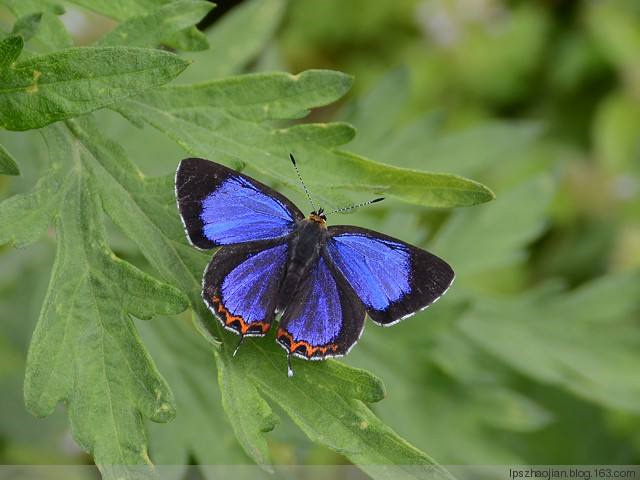}}
  \caption{Butterfly Eco-photograph}
  \label{fig_1}
\end{figure}

The butterfly data sets used in this paper are all photos of butterflies in the natural ecological environment, hereinafter referred to as ecological photos. One part is from the data set provided in document\cite{r14}, the other part is from the images in search engines such as Baidu and image libraries, including 94 species and 11 genera of butterflies. Fig.\ref{fig_1} shows some samples of butterfly ecology.

A total of 5695 pictures were taken from document\cite{r14}, including two kinds of photographs of butterflies: specimen photograph and ecological photograph. According to document\cite{r12}, because the shooting angle and background environment of specimens differ greatly from the ecological photographs, the training effect of using only ecological photographs in the training set is obviously better than that of using both specimens and ecological photographs together in butterfly detection and classification tasks, and the purpose of this study is to locate butterflies in natural environment and determine the species of butterflies, so only 1048 photos of butterflies in natural ecological environment are selected in this paper.

Most of the butterfly samples contained in each photo in the data set are only one, and the maximum number is not more than 20. Each butterfly species consists of at least four samples with a typical heavy-tailed distribution.

The test set is based on the standard test set provided in document\cite{r14}, which contains 678 ecological photos and the rest as training set.

\subsection{Data Annotations}
Because the posture of butterflies in ecological photographs is more complex, and even there are many butterflies overlapping together, and the data sets provided in document\cite{r14} are confused and there is no uniform standard for labeling. We formulated a set of uniform labeling standards and manually labeled the positions and  species of all butterfly samples in all photos according to this standard.

\begin{figure}[htbp]
  \centering
  \subfigure[Uses the antennae and legs of butterflies as the border]{
  \includegraphics[height=3cm]{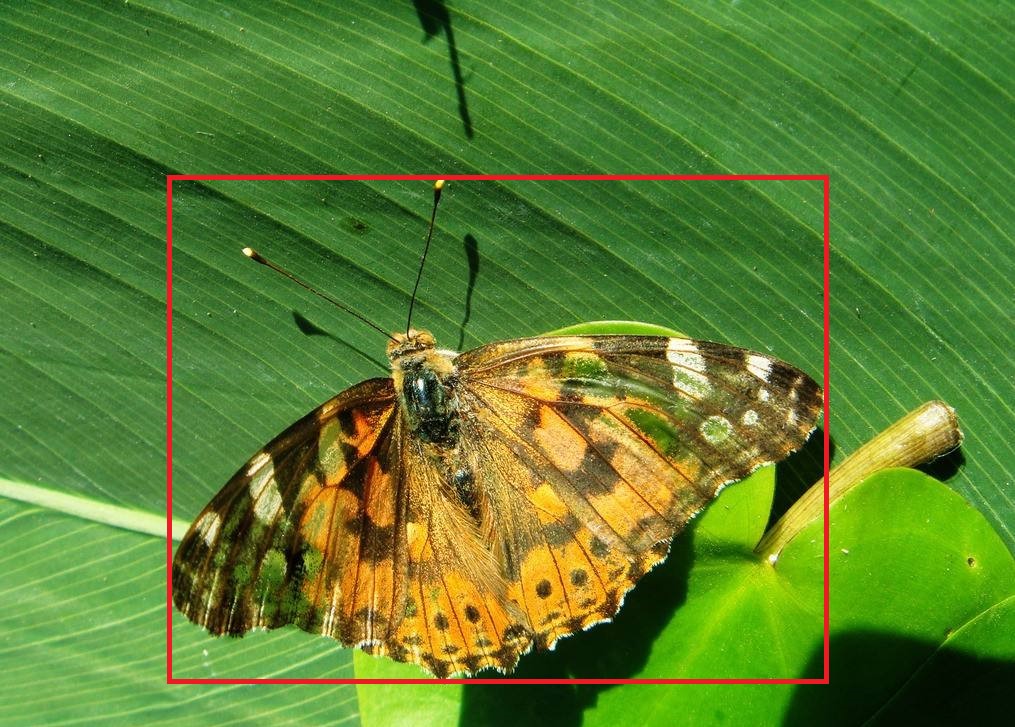}
  \label{fig_2a}
  }
  \quad
  \subfigure[Uses the antennae and legs of butterflies as the border]{
  \includegraphics[height=3cm]{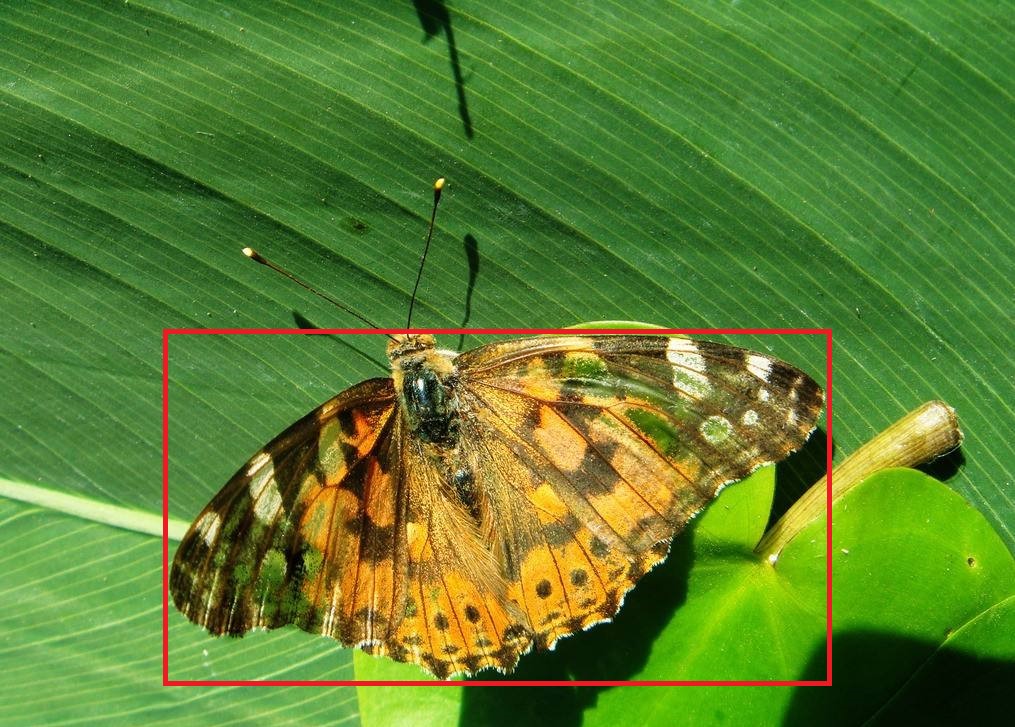}
  \label{fig_2b}
  }
  \caption{Two different methods to annotate a single butterfly}
  \label{fig_2}
\end{figure}

In the data set provided in document\cite{r14}, there are two ways to label the area where butterflies are located: one is to use the antennae and legs of butterflies as the border, as shown in Fig.\ref{fig_2a}; the other is to use the trunk and wings of butterflies as the border, as shown in Fig.\ref{fig_2b}. We use two annotation methods to unify data sets.

\begin{figure}[htbp]
  \centering
  \subfigure[Labels multiple butterflies in overlapping areas as a single sample]{
  \includegraphics[height=3cm]{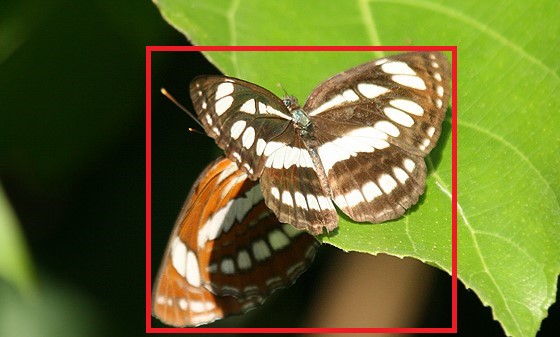}
  \label{fig_3a}
  }
  \quad
  \subfigure[Each butterfly in the overlapping area is independently labeled and the occluded part is ignored]{
  \includegraphics[height=3cm]{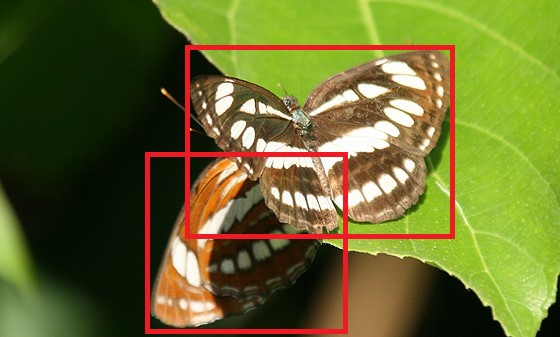}
  \label{fig_3b}
  }
  \caption{Two different methods to annotate two or more overlapping butterflies}
  \label{fig_3}
\end{figure}

Because some butterfly species have social attributes, many butterflies often overlap in photos. The data set provided in document\cite{r14} uses the method of labeling multiple butterflies in overlapping areas as a single sample, as shown in Fig.\ref{fig_3a}. We have also developed a standard for labeling this situation: each butterfly in the overlapping area is independently labeled and the occluded part is ignored, as shown in Fig.\ref{fig_3b}. By using this method, not only the number of training samples is increased, but also the recognition effect of the model for complex scenes is improved.

\subsection{Data Preprocessing}

Target detection algorithms based on deep learning often require a large amount of data as training set. In this paper, we expand the training set by nine transformation methods, such as rotation, mirror image, blur, contrast rise and fall, and combine different pretreatment methods and their parameters (such as rotation angle, exposure, etc.) to get the optimal pretreatment method. The results will be shown in Part 4.

Through the above process, butterfly automatic detection and classification in natural ecological environment has been transformed into a multi-objective detection and classification problem. Different from common target detection problems, butterfly automatic detection and classification problems have three difficulties: 1) There are many classifications (94 classifications); 2) The distribution of samples is not uniform. Some rare species of butterflies have significantly fewer samples than other species of butterflies; 3) It is necessary to classify different small classes (different kinds of butterflies) under the same big class (butterflies), that is, fine-grained classification is needed. Therefore, the research of butterfly automatic detection and classification in this paper is more difficult.

\section{Butterfly detection and Recognition Method}

\subsection{YOLO Model}

YOLO model\cite{r1} proposed by Joe Redmon is a well-known end-to-end learning model in the field of target detection. Its characteristic is that compared with the two-step model of RCNN\cite{r8} series, the execution speed of YOLO model is much faster, and it performs well in fine-grained detection. The third generation model of YOLO, YOLO v3, is chosen in our task.

\begin{figure}
  \centering
  \includegraphics[width=10cm]{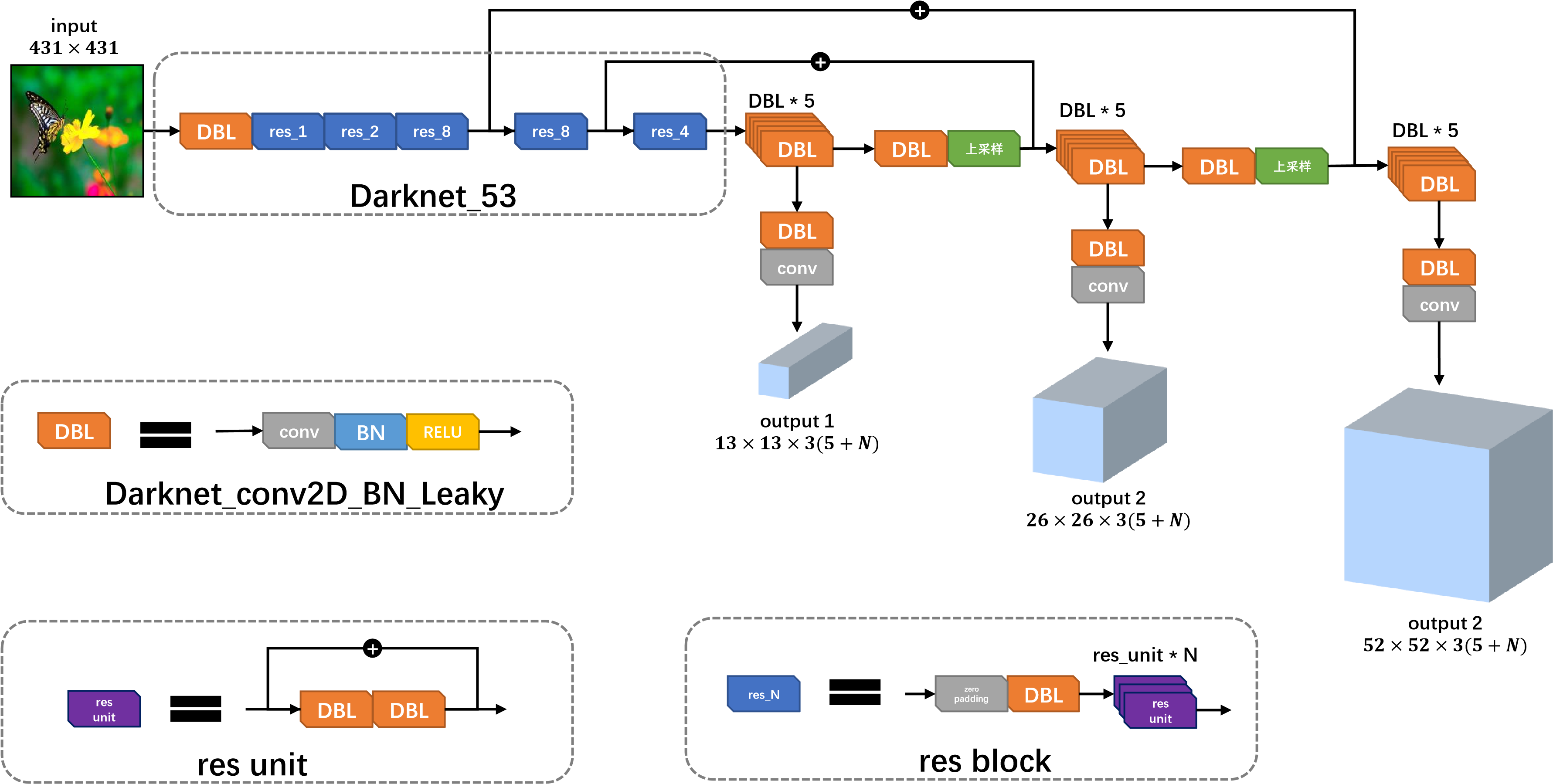}
  \caption{Structure diagram of YOLO V3}
  \label{fig_4}
\end{figure}

The structure of the YOLO V3 model is shown in Fig.\ref{fig_4}. In order to detect butterflies of different sizes (different proportion of area) in natural photographs, YOLO V3 uses multi-scale feature maps to detect objects of different sizes after feature extraction network (darknet-53).

YOLO V3 outputs three feature maps of different scales. After 79 layers of convolution network, the detection results of the first scale are obtained through a three-layer feature extraction network. The feature map used for detection here is 32 times down-sampling of the input image. Because of the high down-sampling ratio, the perceptual field of the feature map is relatively large, so it is suitable for detecting objects with large area in the image.

Up-sampling convolution is done from 79 layers back. The eighty-first layer feature map is joined with the sixty-first layer feature map. After a three-layer feature extraction network, a fine-grained feature map of the ninety-first layer, i.e. a 16-fold down-sampling feature map relative to the input image, is obtained. It has a medium-scale perceptual field of view and is suitable for detecting objects with medium area proportion in image.

Finally, layer 91 feature map is sampled again, and joined with layer 36 feature map. After a three-layer feature extraction network, the feature map with 8 times lower sampling relative to the input image is obtained. It has the smallest perception field and is suitable for detecting objects with small area proportion in the image.

Each output contains $3D^2$ separate $5+N$ dimension vectors, where symbol $D$ represents the edge length of the output feature graph at this scale. Number 3 centers on the number of priori boxes in each grid cell, and symbol N is the number of classifications. The first four dimensions of each vector represent the position of the prediction box, the 5th dimension represents the probability of the target in the candidate box, and the $5 + i$ dimension represents the probability that the target in the candidate box belongs to the category $i$.

YOLO uses mean squares and errors as loss functions. It consists of four parts: prediction box error ($ERR_{center}$), predicting boundary width and height Error ($ERR_{wh}$), classification error ($ERR_{class}$) and prediction confidence error ($ERR_{conf}$) are composed of four parts.

\begin{equation}
  Loss = \lambda_{coord}(ERR_{center}+ERR_{wh})+ERR_{class}+ERR_{conf}
\end{equation}

\begin{equation}
  ERR_{center} = \sum^{D^2}_{i=0}\sum^{n}_{j=0}\iota^{obj}_{ij}[(x_{i}-\hat{x}_{i})^{2}+(y_{i}-\hat{y}_{i})^{2}]
\end{equation}

\begin{equation}
  ERR_{wh} = \sum^{D^2}_{i=0}\sum^{n}_{j=0}\iota^{obj}_{ij}[( \sqrt{w_{i}} - \sqrt{\hat{w}_{i}})^{2}+(\sqrt{h_{i}}-\sqrt{\hat{h}_{i}})^{2}]
\end{equation}

Here, $n$ is the number of prediction frames in a grid cell, $(x, y)$ is the center coordinate of the prediction frame, and $w$ and $h$ are the width and height of the prediction frame, respectively. Because the error caused by large prediction box is obviously larger than that caused by small prediction box, YOLO adopts the method of predicting square root of width and height instead of directly predicting width and height. If the jth prediction box in the 1st grid cell is responsible for the object, then we have $\iota_{ij}^{obj}=1$, conversely, $\iota_{ij}^{obj}=0$.

\begin{equation}
  ERR_{class} = \sum^{D^2}_{i=0}\iota^{obj}_{ij}\sum^{n}_{j=0}[p_i(c) - \hat{p_i}(c)]^2
\end{equation}

YOLO considers that each grid cell contains only one classified object. If $c$ is the correct category, then $\hat{p_i}(c)=1$, conversely, $\hat{p_i}(c)=0$.

\begin{equation}
  ERR_{conf} = \sum^{D^2}_{i=0}\sum^{n}_{j=0}\iota^{obj}_{ij}(c_i-\hat{c_i})^2 + \lambda_{noobj}\sum^{D^2}_{i=0}\sum^{n}_{j=0}(1-\iota^{obj}_{ij})(c_i-\hat{c_i})^2
\end{equation}

where $c_i$ denotes the confidence of objects contained in the prediction box. If there is an object in the real boundary box, $\hat{C_i}$ is the IoU value of the real boundary box and the prediction box. Conversely, there are $\hat{C_i}=0$.

The parameter $\lambda$ is used in different weighted parts of the loss function to improve the robustness of the model. In this paper, we have $\lambda_{coord}=5$, $\lambda_{coord}=0.5$.

\subsection{Integrated YOLO algorithm}

In order to get more accurate classification and detection results and improve the generalization ability of magic, this paper further processes the results of multiple YOLO models, and obtains the integrated YOLO algorithm. The pseudo-code of the algorithm is shown in Algorithm.\ref{alg_1}.

\begin{figure}
  \centering
  \includegraphics[width=10cm]{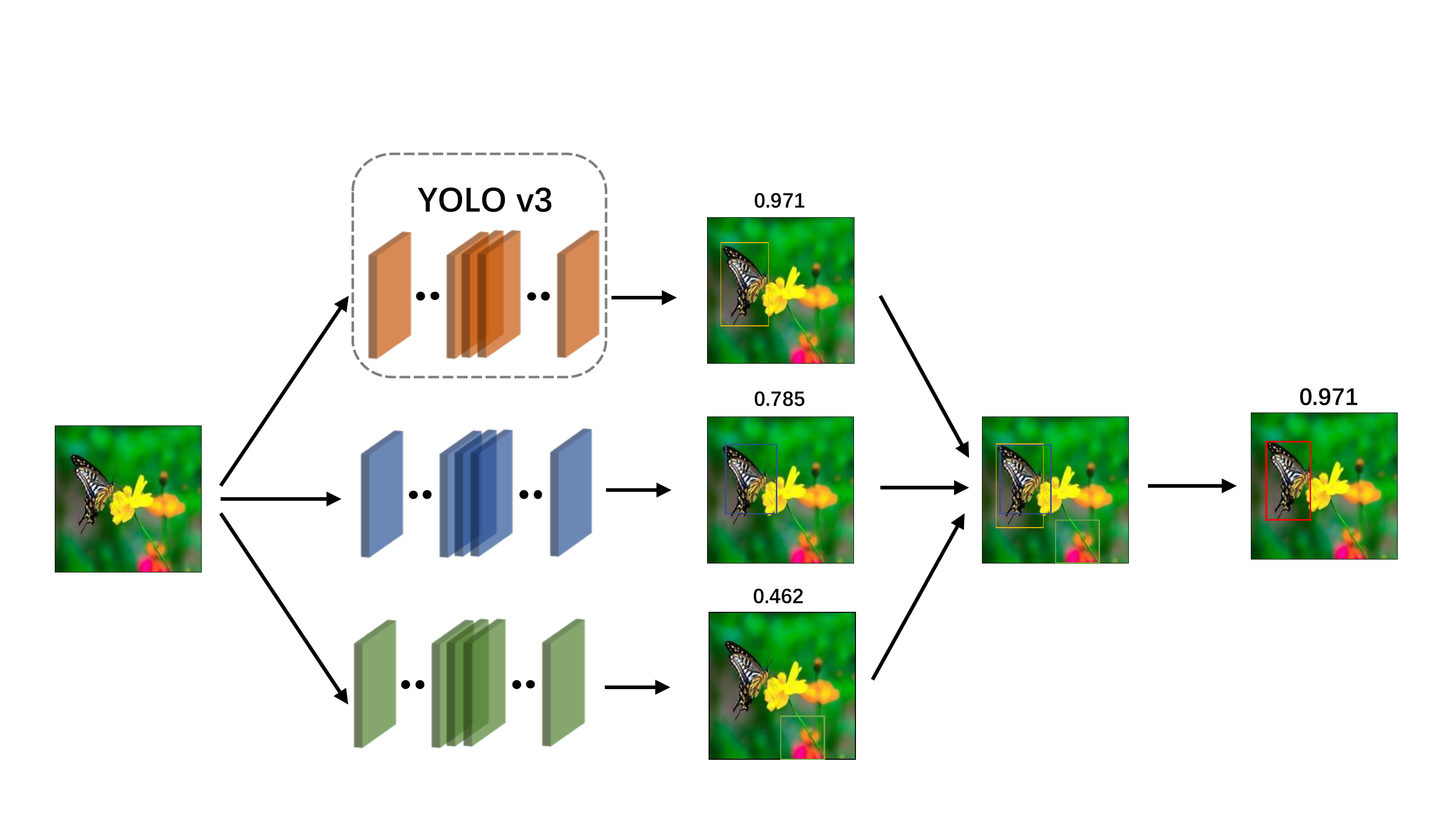}
  \caption{Structural diagram of integrated YOLO algorithm}
  \label{fig_5}
\end{figure}

Its core idea is to use multiple models with better training effect to predict the pictures separately and cluster the prediction frames. The clustering process is shown in Fig.\ref{fig_5}.

Each positioning box can be described as a four-dimensional vector $b_{i}= (x_{1i},x_{2i},y_{1i},y_{2i})$, an integer $c_{i}$ and a real number $p_{i}$. It represents: in a rectangle with upper left corner $x_{1i}, y_{1i}$, lower right corner $x_{2i}$ and $y_{2i}$, the probability of having a butterfly with the species number $c_{i}$ is $p_{i}$.

Let each cluster set be $S_{1},S_{2},...,S_{k}$. For each $S_{i} (i = 1, 2,..., k)$, satisfy: for $\forall i, j \in S,c_{i}=c_{j}$. Define the "summary" of a set as the following:

\begin{equation}
  \left\{
  \begin{array}{rl}
    B(S) & = \frac{1}{\sum_{i \in S}P_{i}}\sum_{i \in S}P_{i}b_{i}
    \\
    P(S) & =\frac{1}{\mid S \mid}\max_{i \in S}P_{i}
    \\
    C(S) & =c_{i} (i \in S)
  \end{array}
  \right.
\end{equation}

where, $B(S)$ is the "aggregate" positioning box of set $S$, $P(S)$ is the "aggregate" probability of set $S$, and $C(S)$ is the "aggregate" classification of set $S$. Each time a single predicted bounding box is categorized, the set $S$ with the highest probability $P(S)$ is selected from all sets with the same classification as the detection box and $IoU(B(S),b) \ge 0.5$. If there is no $S$ that meets the criteria, place the box in a new set $S_{k+1}$

\begin{algorithm}[htb]
  \label{alg_1}
  \caption{Merge Boxes}
  \begin{algorithmic}[1]
    \Require
    $n$: number of boxes;
    $b$: boxes;
    $c$: classes;
    $p$: probability;
    \Ensure
    $S$: Merged boxes, sorted by P(S);
    \State S $\gets$ $\emptyset$
    \State k $\gets$ 0 \\
    \For{$i=1,2,\cdots,n$}
    {
      \For{$j=1,2,\cdots,k$}
      {
        \If{$C(S_i)=C_i$ and $IoU(b_i,b(S_j) > 0.5)$}
        {
          Insert $b_i$ to $S_j$
        }
      }
      \If{$\not\exists S_j (j=0,1,\cdots,k), st. i \in S_j$}
      {
        k $\gets$ k + 1; $S_k$ $\gets$ ${b_i}$
      }
      Sort S by $P(S_i)$
    }
    \State
    \Return $B(S_i), P(S_i), C(S_i) (i=1, 2, \cdots, k)$;
  \end{algorithmic}
\end{algorithm}

\section{Experimental results and analysis}

\subsection{Evaluation Index}

In this paper, the intersection over union(IoU) is used as the evaluation index of butterfly positioning task, which is defined as the ratio of the area of two regions to the area of merging. The optimal ratio is 1, i.e. complete overlap. In the experiment, $IoU=0.5$ is taken as the threshold value, i.e. the prediction box and the original tag box are positively positioned with $IoU > 0.5$, and the $IoU \le 0.5$ is positively positioned with error.
In this paper, the mean average precisio·n mean (mAP) is used as the evaluation index of butterfly classification task. mAP is derived from precision (pre) and recall (recall). The calculation formula is as follows.

\begin{equation}
  pre = \frac{TP}{TP + FP}
\end{equation}

\begin{equation}
  recall = \frac{TP}{TP + FN}
\end{equation}

where TP(rue-positive number), FP(false-positive number) and FN(false-negative number) represent the number of positive samples predicted to be positive, the number of negative samples predicted to be positive and the number of positive samples predicted to be negative, respectively.

According to different confidence levels, we can get several (pre, recall) points, and draw the pre-recall curve with recall rate as the horizontal coordinate and precision rate as the longitudinal coordinate. Where, the average precision AP is the area around the pre-recall curve and the recall axis, which is the integral of the pre-recall curve, as shown below

\begin{equation}
  AP = \int^1_0 p(r)dr
\end{equation}

In practical applications, the sum of rectangular areas is generally used to approximate the integral. In this paper, we use the PASCAL VOC Challenge's calculation method after 2010\cite{r15}, i.e. the recall can be divided into n blocks $[0,\frac{1}{n},...,\frac{(n-1)}{n},1]$, then the average accuracy (AP) can be expressed as

\begin{equation}
  AP = \frac{1}{n}\sum^{n}_{i=1}max_{r \in [\frac{i-1}{n},\frac{i}{n}]}p(r)
\end{equation}

The mean average precision(mAP) of all classes can be expressed as:

\begin{equation}
  mAP = \frac{1}{n}\sum^{N}_{i=1}AP(n)
\end{equation}

\subsection{YOLO V3 model effect experiment}

In this paper, two different methods of butterfly labeling are tested: one is to use the antennae of butterflies as the boundary of the frame, which is called the full-scale labeling method; the other is to use the trunk and wings of butterflies as the boundary of the frame, which is called the non-full-scale labeling method. Taking different preprocessing methods, the best results of the two annotation methods are shown in Table 1.

\begin{table}[t]
  \begin{center}
    \begin{tabular}{|c|c|}
      \hline
      \textbf{Annotation Method}         & \textbf{mAP} \\
      \hline
      the full-scale labeling method     & $0.734$       \\
      \hline
      the non-full-scale labeling method & $0.777$       \\
      \hline
    \end{tabular}
  \end{center}
  \caption{Under different annotations, the best mAP for a single YOLO v3 model on a butterfly test set.}
  \label{tab:1}
\end{table}

It can be seen that the results obtained by the non-full-scale method are obviously better than that by the full-scale method. Because the area around the antennae of butterflies is larger in the full-scale method, this can make the proportion of the area around the antennae of butterflies in the labeled area decrease, thus the influence of background environment on classification can become larger. The non-full labeling method is more suitable for butterfly automatic detection and recognition tasks.

Because the different size of the input image in YOLO model will lead to the different number of mesh cells in different output scales, we test the performance of single YOLO V3 model in butterfly automatic localization task under different input sizes in this paper.

Using the non-full-scale method, without any other pretreatment, the best results of the three input sizes are shown in Table 2.

\begin{table}[t]
  \begin{center}
    \begin{tabular}{|c|c|}
      \hline
      \textbf{Image Size (px $\times$ px)} & \textbf{mAP} \\
      \hline
      416 $\times$ 416                     & $0.475$        \\ \hline
      608 $\times$ 608                     & $0.553$        \\ \hline
      736 $\times$ 736                     & $0.447$        \\ \hline
    \end{tabular}
  \end{center}
  \caption{The mAP of the YOLO V3 model in the butterfly test set when different sizes of images are used as input.}
  \label{tab:2}
\end{table}

It can be seen that the resolution input using 608px $\times$ 608px achieves better accuracy.

we also test the performance of a single YOLO V3 model in butterfly automatic positioning and classification tasks when different pretreatment methods are used to label 608px $\times$ 608px input with the non-full labeling method. Table 3 shows the effects of two parameters that have a greater impact on the results of classification.

\begin{table}[t]
  \begin{center}
    \begin{tabular}{|c|c|c|}
      \hline
      \textbf{Rotations}                                                               & \textbf{Saturation}           & \textbf{mAP}   \\ \hline
      Null                                                                             & NULL                          & 0.553          \\ \hline
      (0\degree, 45\degree, 90\degree, 180\degree)                                     & NULL                          & \textbf{0.691} \\ \hline
      (0\degree, 45\degree, 90\degree, 135\degree, 180\degree, 255\degree, 270\degree) & NULL                          & 0.681          \\ \hline
      (0\degree, 45\degree, 90\degree, 180\degree)                                     & (1.0,1.5)                     & 0.0691         \\ \hline
      (0\degree, 45\degree, 90\degree, 180\degree)                                     & (1.0,1.2,1.5,1.8)             & \textbf{0.777} \\ \hline
      (0\degree, 45\degree, 90\degree, 180\degree)                                     & (1.0,1.3,1.5,1.7)             & \textbf{0.753} \\ \hline
      (0\degree, 45\degree, 90\degree, 180\degree)                                     & (1.0,1.2,1.3,1.5,1.7,1.8,2.0) & \textbf{0.723} \\ \hline
    \end{tabular}
  \end{center}
  \caption{mAP of YOLO v3 model on butterfly test set under different and processing modes.}
  \label{tab:3}
\end{table}

It can be seen that the best classification results are obtained by rotating the original images (0\degree, 45\degree, 90\degree, 180\degree) and exposing them to 1.0, 1.2, 1.5 and 1.8 times respectively. The value of mAP can be reached to 0.7766. In particular, the recognition rate of butterflies with dark and protective colors is significantly higher than that of the first group.

\subsection{Integrated Model Effectiveness Experiments}

We chose the three best performing models in a single YOLO model (Models 3, 4, 5 in Table 3) for integration. In detection task, 98.35\% accuracy is obtained. In classification task, 0.798 mAP is obtained in the classification of test sets (94 classes) and 0.850 mAP is obtained in the classification of families (11 classes).

Table 4 shows the performance of the integrated Yolo model and the mainstream target detection models, such as fast-RCNN, YOLO v3 etc., in butterfly automatic detection and classification tasks.

\begin{table}[t]
  \begin{center}
    \begin{tabular}{|c|c|}
      \hline
      \textbf{Model}               & \textbf{mAP}   \\ \hline
      Faster-RCNN + ZF             & 0.733          \\ \hline
      Faster-RCNN + VGG CNN M 1024 & 0.726          \\ \hline
      Faster-RCNN + VGG16          & 0.761          \\ \hline
      YOLO v3                      & 0.777          \\ \hline
      Integrated YOLO              & \textbf{0.798} \\ \hline
    \end{tabular}
  \end{center}
  \caption{mAP results on butterfly test sets under different models.}
  \label{tab:4}
\end{table}

The above results show that the data annotation and preprocessing methods presented in this paper are suitable for butterfly automatic detection and classification tasks. It also shows that the integrated YOLO algorithm proposed by us is very effective and correct for butterfly detection and species identification in natural ecological photos.

\section{Summary and Prospect}

At present, the field of target detection is mainly divided into two schools: end-to-end detection and distributed detection. End-to-end detection is very fast, but there is a big gap between the accuracy and the distributed detection scheme. Taking advantage of the fast speed of the end-to-end model and the high similarity between butterfly species, we propose an integrated model based on YOLO with different concentrations, which maintains the detection speed of the end-to-end model and improves the detection accuracy and positioning accuracy.

The essence of the integrated model is to find a better solution based on the optimal solution of a model under various conditions. By this way, the performance of the model on a specific training set can be improved, and the comprehensive generalization ability of the model can be improved. In the test set provided in reference\cite{r14}, the model achieves 98.35\% accuracy in the task of locating ecological photos, 0.7978 map in the task of locating and identifying species, and 0.8501 map in the task of locating and identifying subjects.

The fact that butterflies have a high similarity among species indicates that there is a strong relationship among all kinds of butterflies in the knowledge map. For a wider range of target detection tasks, species can be accurately classified and located by means of knowledge maps, and their recognition tasks can be further optimized by means of different model capabilities and different recognition focus.

\section{Acknowledgement}

Thank you here for the organizers of Baidu Encyclopedia Butterfly related pages, collectors of university open biological data sets and collectors of data provided by China Data Mining Conference. They have provided valuable data for our experiments.

%
%

\end{document}